\documentclass[conference]{IEEEtran}

\usepackage{cite}
\usepackage{amsmath,amssymb,amsfonts}
\usepackage{algorithmic}
\usepackage{graphicx}
\usepackage{textcomp}
\usepackage{xcolor}
\usepackage{booktabs}
\usepackage{float}

\def\BibTeX{{\rm B\kern-.05em{\sc i\kern-.025em b}\kern-.08em
    T\kern-.1667em\lower.7ex\hbox{E}\kern-.125emX}}
    
\begin{document}

\title{Depth-Aware Pothole Detection Using YOLO and RT-DETR at the Edge}

\author{
    \IEEEauthorblockN{Md Monjurul Ahsan Prodhan, Md Nour Hossain}
    \IEEEauthorblockA{\textit{College of Emergency Preparedness, Homeland Security and Cybersecurity} \\
    \textit{University at Albany, State University of New York} \\
    Albany, United States \\
    mprodhan@albany.edu, mhossain6@albany.edu}
}

\maketitle

\begin{abstract}
Pothole detection and its severity measurement is still an important challenges in urban infrastructure management, where late maintenance directly contributes to vehicle damage, road accidents, and escalating repair costs. Existing automated approaches depend on 2D RGB images and cannot measure physical depth of potholes. In this paper, we present a depth-aware pothole detection framework and then compare five architectures: YOLOv8n, YOLOv8n-Seg, YOLOv9t, RT-DETR-L, and RT-DETR-X for RGB-D sensor fusion-based detection and automated depth measurement. A custom offline augmentation pipeline is used here to simulate adverse road monitoring conditions. All models are trained on the PothRGBD dataset with an 80\% training and 20\% validation split and evaluated using Precision, Recall, mAP@50, and mAP@50-95. Before measuring the depth data, all depth maps are corrected for camera tilt using RANSAC ground-plane orthorectification and all zero-valued sensor pixels are cast to NaN before any statistic is computed. YOLOv8n-Seg achieves the highest mAP@50 of 0.9556 and mAP@50-95 of 0.6758 with the most accurate depth estimate of 2.96 cm with the pixel-precise Dseg algorithm. YOLOv8n achieves the fastest inference at 3.6ms. RT-DETR-X achieves the highest detection confidence at 92.70\%. An important finding is that even after full RANSAC orthorectification, bounding box models overestimate pothole depth by 0.16–0.21 cm compared to pixel-precise segmentation masks. This confirms that the pavement inclusion bias is structural rather than a calibration artifact.
\end{abstract}

\begin{IEEEkeywords}
Pothole Detection, RGB-D Sensing, Convolutional Neural Networks, Transformer Networks
\end{IEEEkeywords}

\section{Introduction}
Road infrastructure deterioration, mainly the formation of potholes, represents one of the most persistent and economically big challenges facing municipal governments worldwide. Potholes are caused by heavy traffic combined with water seepage and freezing temperatures that break down the road over time. The scale of this problem is severe and growing. In the United States alone, an estimated 55 million potholes exist across the country's nearly four million miles of roads, with more than 30\% of urban roads and highways classified as being in mediocre condition~\cite{b28}. In the United States alone, potholes contribute to over 5,000 traffic accidents annually: mechanically, striking a severe road cavity generates impact forces equivalent to a 35-mph vehicle collision~\cite{b29}. Motorcyclists and cyclists face disproportionate risk, with surveys indicating that 31\% of cyclists have been involved in accidents or similar cases caused directly by poor road surfaces~\cite{b29}. Beyond the United States, the global situation is also alarming. In the United Kingdom, 451 people were killed or injured due to potholes between 2018 and 2022 and local authorities paid out over \pounds32 million in personal injury compensation over the same period~\cite{b30}. In India, urban roads report over 2.5 million potholes during monsoon season~\cite{b28}. In the United Kingdom, the Centre for Economics and Business Research estimated the total economic damage from potholes at \pounds14.4 billion per year in England alone~\cite{b30}.

Traditional inspection depends on manual crew surveys that are labor-intensive and unable to provide quantitative cavity depth measurements the most critical indicator for repair prioritization. Potholes are normally identified only after causing damage rather than being graded proactively based on objective severity.

To address these limitations, this paper presents a depth-aware pothole detection framework conducting a comprehensive ablation study comparing five state-of-the-art architectures: YOLOv8n (Bounding Box) (YOLO = You Only Look Once), YOLOv8n-Seg (Instance Segmentation), YOLOv9t, RT-DETR-L (RT-DETR = Real-Time Detection Transformer), and RT-DETR-X under a unified experimental protocol on the PothRGBD dataset~\cite{yurdakul2025enhanced}. A custom offline augmentation pipeline is constructed using the Albumentations library~\cite{buslaev2020albumentations}. This is designed to simulate the harsh visual conditions encountered during real-world municipal road monitoring. A RANSAC ground-plane orthorectification step corrects camera tilt before any depth statistic is computed.

The key contributions of this work are as follows. First, we conduct a systematic empirical ablation study comparing CNN and Real-Time Transformer architectures evaluated specifically for RGB-D physical depth extraction accuracy in a pothole severity assessment context. Second, we design and validate a domain-specific adverse-condition augmentation pipeline comprising six targeted Albumentations techniques. Third, a RANSAC ground-plane orthorectification step that enables physically meaningful depth measurement from an angled sensor. Fourth, we quantify and expose the spatial measurement bias. This paper is organized as follows: Section~\ref{sec:related} reviews related work, Section~\ref{sec:methodology} details the proposed methodology, Section~\ref{sec:experiments} presents experimental results and analysis, and Section~\ref{sec:conclusion} concludes with directions for future research.

\section{Related Works}

\label{sec:related}

Research on pothole detection has evolved over the past decade. This section reviews the most relevant literature in these areas: CNN-based, Transformer architectures based, RGB-D sensor fusion based, and image augmentation strategies for adverse condition robustness.

\subsection{CNN-Based Pothole Detection}

Previous work on pothole detection mostly dependent on manual features such as edge detection, texture analysis, and threshold-based segmentation on grayscale or RGB images~\cite{varona2020deep}. These methods were fast but struggled to generalize across different road surfaces, lighting conditions, and camera setups. The shift to deep convolutional neural networks changed this significantly, as models could now learn features directly from raw image data end-to-end, without manual feature engineering.

Redmon et al. introduced the YOLO architecture as a unified, single-stage object detection framework capable of real-time inference by reformulating detection as a single regression problem over a grid of spatial predictions~\cite{redmon2016you}. The single-stage design jointly predicts bounding box coordinates and class probabilities in a single forward pass, making it substantially faster than two-stage detectors such as Faster R-CNN~\cite{ren2015faster} while maintaining competitive accuracy. YOLOv9 introduced Programmable Gradient Information (PGI) and the Generalized Efficient Layer Aggregation Network (GELAN)~\cite{wang2024yolov9}.

Maeda et al.~\cite{maeda2018road} explained that YOLO-based architectures could achieve strong road damage detection accuracy using only smartphone cameras. Naddaf-Sh et al.~\cite{naddaf2020efficient} applied deep convolutional networks to pothole detection under varying environmental conditions.

He et al.~\cite{he2017mask} introduced Mask R-CNN, demonstrating that pixel-precise instance masks could be generated with modest additional computational overhead over a standard object detector. YOLOv8-Seg integrates this capability into the single-stage YOLO framework, producing instance masks that conform precisely to detected object boundaries. In the context of pothole depth extraction, pixel-precise masks ensure that depth measurements are computed exclusively over the cavity region.

\subsection{Transformer Architectures for Object Detection}
The application of self-attention mechanisms to computer vision tasks has produced a new generation of object detection architectures with fundamentally different inductive biases compared to convolutional networks. Carion et al. introduced DETR (Detection Transformer)~\cite{carion2020end}, which reformulated object detection as a direct set prediction problem using a Transformer encoder-decoder architecture and bipartite matching loss. By eliminating hand-crafted components such as anchor generation and Non-Maximum Suppression (NMS), DETR offered a cleaner end-to-end detection pipeline~\cite{zhao2024detrs}.

Zhao et al. further advanced this direction with RT-DETR (Real-Time Detection Transformer)~\cite{zhao2024detrs}, which employs an efficient hybrid encoder combining intra-scale feature interaction and cross-scale feature fusion, enabling inference speeds competitive with leading YOLO models while retaining the global reasoning capabilities of attention mechanisms. RT-DETR additionally eliminates the NMS post-processing step entirely, removing a significant source of latency and hyperparameter sensitivity that affects all YOLO-based architectures. The self-attention mechanism in RT-DETR gives the model to capture long-range spatial dependencies across an entire image in a single operation, making it particularly well-suited for detecting irregular, low-contrast anomalies such as potholes under adverse visual conditions where local convolutional features may be insufficient~\cite{zhao2024detrs}.

\subsection{RGB-D Sensor Fusion for Depth Estimation}
RGB-D sensors, which simultaneously capture aligned color and depth information, offer a practical and cost-effective alternative to high-precision LiDAR systems for infrastructure monitoring applications. Consumer-grade structured-light sensors such as the Intel RealSense D415 can provide per-pixel depth measurements at ranges suitable for road surface analysis at a fraction of the cost of LiDAR-based surveying equipment~\cite{henry2012rgb}. Current works have explored fusing 2D detector outputs with aligned depth maps to estimate pothole physical properties from consumer-grade RGB-D sensors~\cite{huang2023deep}.

\subsection{Image Augmentation for Adverse Condition Robustness}

Standard augmentation strategies such as random cropping, horizontal flipping, color jittering, and mosaic composition are commonly applied during training and are natively supported by frameworks such as Ultralytics~\cite{jocher2023ultralytics}. However, for road monitoring applications, these generic augmentations are not enough to capture the domain-specific visual degradations~\cite{hendrycks2019benchmarking}. In the autonomous driving domain, Tremblay et al.~\cite{tremblay2018training} demonstrated that combining synthetic weather augmentations with real training data substantially improved object detection robustness under adverse conditions, validating the approach adopted in this work.

\subsection{RANSAC Orthorectification}
The RANSAC (Random Sample Consensus) algorithm has been widely used in computer vision for plane estimation and ground surface fitting in 3D point clouds since it was first presented by Fischler and Bolles \cite{fischler1981random} as a reliable technique for fitting mathematical models in the presence of outliers. When it comes to pothole depth extraction, the camera's physical pitch angle with respect to the road surface introduces a systematic tilt component into raw depth maps obtained from a vehicle-mounted RGB-D sensor. If this tilt is not corrected, any depth statistic calculated within a detection region reflects the camera orientation as much as the actual cavity geometry.

\section{Methods}
\label{sec:methodology}

\begin{figure}[h]
    \centering
    \includegraphics[width=.7\columnwidth]{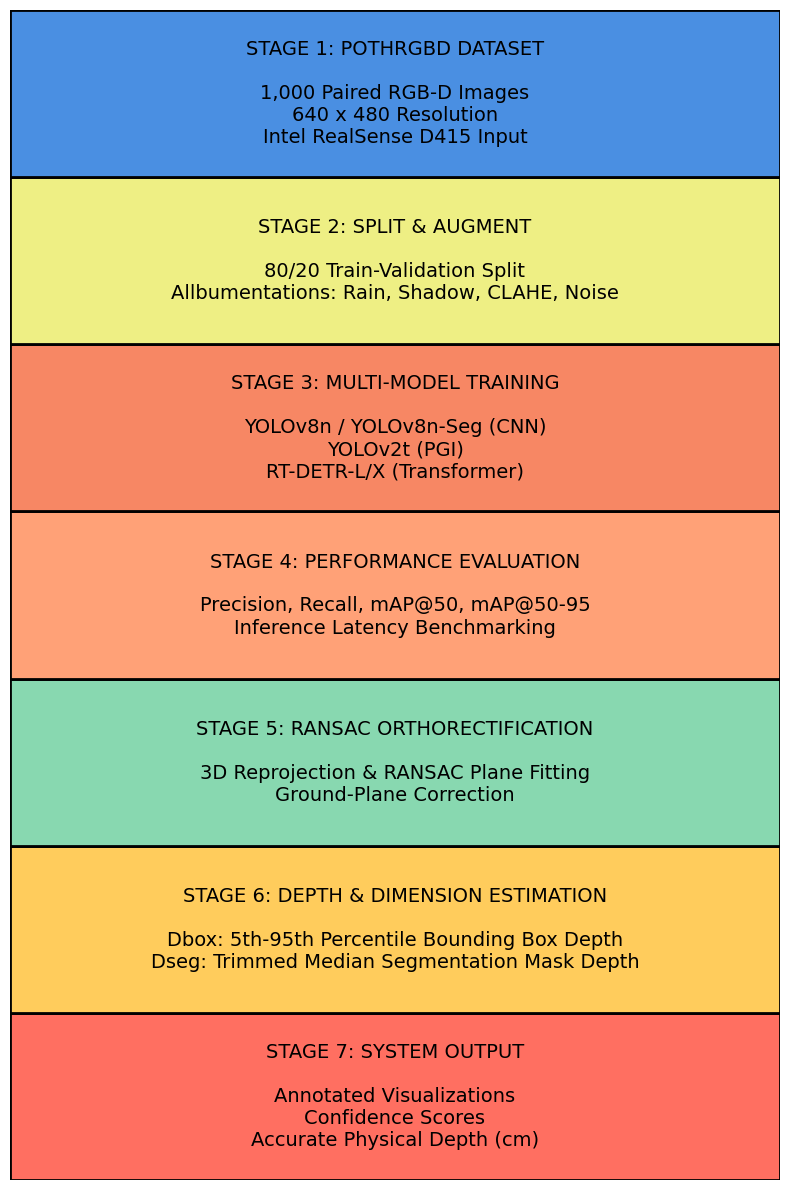}
    \caption{Pipeline of the depth-aware pothole detection and severity assessment.}
    \label{fig:pipeline}
\end{figure}

This section explains the proposed pipeline for RGB-D sensor fusion-based pothole detection and physical depth extraction shown in figure~\ref{fig:pipeline}.

\subsection{Dataset and Splitting}

All experiments are conducted on the PothRGBD dataset~\cite{yurdakul2025enhanced}, comprising 1,000 synchronized RGB-depth image pairs of real-world potholes collected using the Intel RealSense D415 depth camera at $640 \times 480$ pixel resolution. Each RGB image is paired with an aligned depth map in NumPy format (.npy) providing per-pixel distance measurements in millimeters. A fixed random seed of 42 is applied before shuffling, and the dataset is partitioned into an 80\% training split (800 images) and a 20\% validation split (200 images).

\subsection{Augmentation Pipeline}

A domain-specific augmentation pipeline is built using Albumentations~\cite{buslaev2020albumentations} to simulate adverse conditions encountered during real-world road monitoring. Six techniques are applied, each at $p = 0.4$: \textbf{RandomRain} simulates precipitation and surface reflections; \textbf{RandomShadow} replicates shadows from roadside structures; \textbf{CLAHE} (clip 4.0, grid $8{\times}8$) simulates headlight glare on wet road; \textbf{RandomGamma} (80--120) replicates low-light attenuation; \textbf{ISONoise} (shift 0.01--0.05, intensity 0.1--0.5) simulates sensor grain; and \textbf{MultiplicativeNoise} (0.9--1.1, element-wise) introduces transmission artifacts.

\begin{figure}[h]
    \centering
    \includegraphics[width=0.99\columnwidth]
    {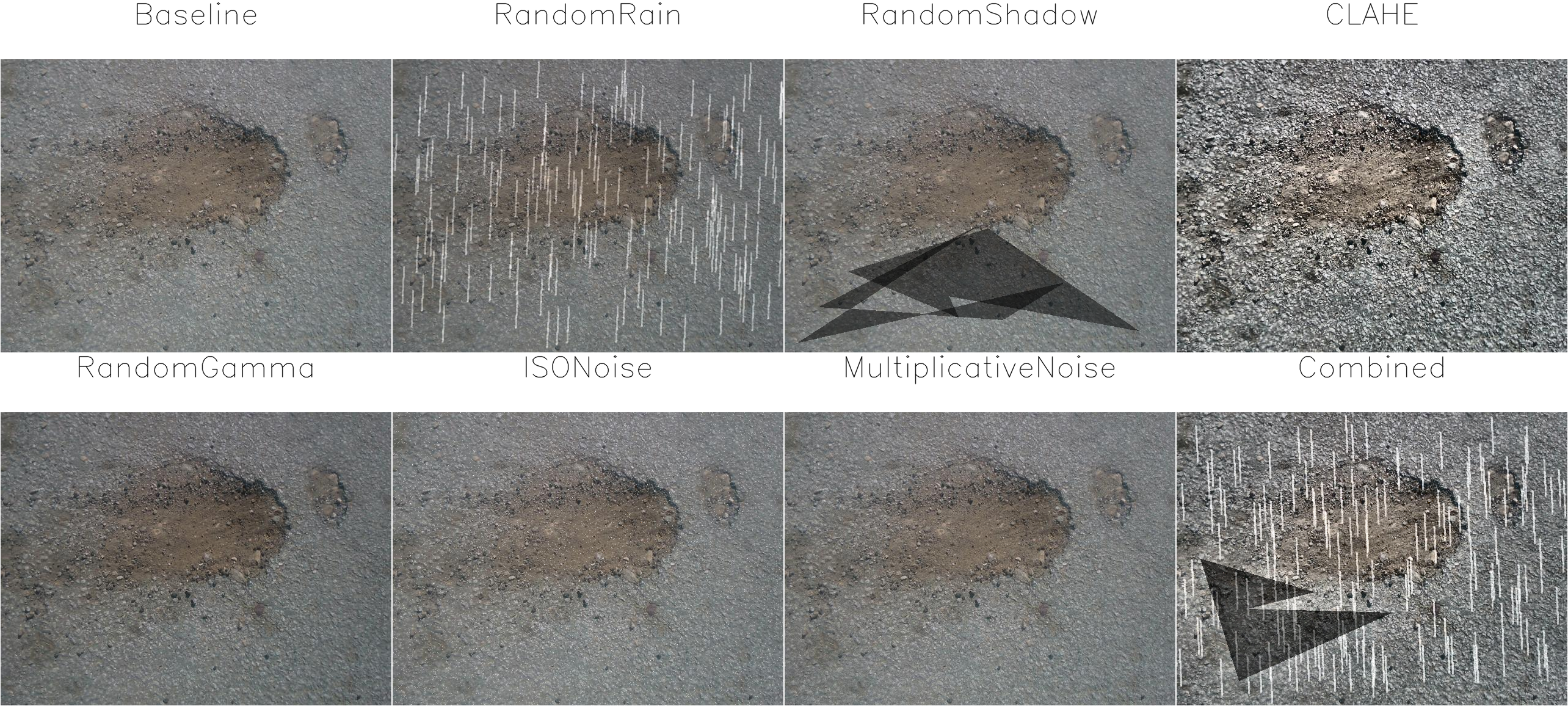}
    \caption{\small Example of
    the domain-specific augmentation
    pipeline applied to a single pothole image from the PothRGBD dataset.}
    \label{fig:augmentation}
\end{figure}

Figure~\ref{fig:augmentation} shows a random image of training data in augmentation pipeline.

\subsection{Architecture-Aware Model Training}
Five architectures are evaluated under a unified training protocol: YOLOv8n~\cite{jocher2023ultralytics}, a lightweight CNN-based bounding-box detector with 3.01M parameters and 8.1 GFLOPs; YOLOv8n-Seg~\cite{he2017mask}, which extends YOLOv8n with pixel-precise instance segmentation and has 3.26M parameters and 12.6 GFLOPs; YOLOv9t~\cite{wang2024yolov9}, which incorporates PGI and the GELAN and has 2.00M parameters and 7.9 GFLOPs; RT-DETR-L~\cite{zhao2024detrs}, a real-time detection transformer with a hybrid encoder that combines convolutional feature extraction and global self-attention, with 31.99M parameters and 103.4 GFLOPs; and RT-DETR-X~\cite{wu2025improved}, shown in Table~\ref{tab:training_config}.

\begin{table}[htbp]
\caption{Training Configuration for All Five Architectures}
\label{tab:training_config}
\centering
\renewcommand{\arraystretch}{1.2}
\begin{tabular}{|l|c|c|c|c|c|}
\hline
\textbf{Param} & \textbf{v8n} & \textbf{v8n-Seg} & \textbf{v9t} & \textbf{DETR-L} & \textbf{DETR-X} \\
\hline
Epochs  & 50   & 50   & 50   & 50     & 50     \\
\hline
ImgSz   & 640  & 640  & 640  & 640    & 640    \\
\hline
Batch   & 16   & 16   & 16   & 8      & 8      \\
\hline
Optim   & Auto & Auto & Auto & AdamW  & AdamW  \\
\hline
$lr_0$  & Auto & Auto & Auto & 0.0001 & 0.0001 \\
\hline
Mosaic  & 0.0  & 0.0  & 0.0  & 0.0    & 0.0    \\
\hline
Mixup   & 0.0  & 0.0  & 0.0  & 0.0    & 0.0    \\
\hline
\end{tabular}
\end{table}

\subsection{Ground-Plane Orthorectification}
A ground-plane correction step was applied before measuring depth. The correction proceeds in four stages. First, we clean the raw depth map by casting all zero-valued pixels to NaN. The Intel RealSense D415 outputs zero for dead pixels and out-of-range measurements, and leaving these in would corrupt any subsequent statistical calculation. Second, we reproject the cleaned 2D depth map into a 3D point cloud using the D415's approximate factory intrinsics ($f_x = f_y = 597.5$ pixels), with the principal point derived from the image center. The standard pinhole camera model governs this reprojection:

\begin{equation}
X = \frac{(u - c_x) \cdot Z}{f_x}, \quad Y = \frac{(v - c_y) \cdot Z}{f_y}, \quad Z = \text{depth}
\end{equation}

\noindent where $(u, v)$ are pixel coordinates and $(c_x, c_y)$ is the principal point. Points with NaN depth values or distances beyond 3{,}000~mm are discarded as sensor noise.

Third, we fit a mathematical plane to the dominant road surface in this point cloud using RANSAC~\cite{fischler1981random} with a residual threshold of 20~mm and a fixed random state of 42 for reproducibility. The fitted plane takes the form $Z = aX + bY + d$, from which we extract the surface normal vector $\mathbf{n} = [-a,\; -b,\; 1]^\top$ and normalize it to unit length. We then compute a rotation matrix $\mathbf{R}$ using the Rodrigues formula to align $\mathbf{n}$ with the world vertical axis $[0,\; 0,\; 1]^\top$. Applying this rotation to every point in the cloud effectively removes the camera's physical pitch and roll, flattening the road surface so that it lies perpendicular to the Z-axis.

Fourth, we project the corrected 3D point cloud back onto a regular 2D grid that matches the original image resolution, using linear interpolation to fill the grid cells. The result is an orthorectified depth map in which depth differences between any two points represent true vertical distances, independent of how the camera was mounted.

\subsection{RGB-D Physical Depth Extraction}
Once the orthorectified depth map is ready, we fuse it with each model's detection output to estimate the physical depth of the detected pothole. As bounding box models and segmentation models create geometrically different outputs, we use two separate depth extraction algorithms.

\subsubsection{Bounding Box Depth (YOLOv8n, YOLOv9t, RT-DETR-L, RT-DETR-X)}

For models that output a rectangular bounding box $(x_1, y_1, x_2, y_2)$, we crop the orthorectified depth map to that region and collect all valid (non-NaN) depth values. Rather than using the raw minimum and maximum, which are easily corrupted by a single dead pixel or sensor spike, we compute depth using robust percentile statistics:

\begin{equation}
D_{\text{box}} = \frac{P_{95} - P_{5}}{10}
\label{eq:dbox}
\end{equation}

\noindent where $P_{95}$ and $P_{5}$ are the 95th and 5th percentiles of the valid depth values (in millimeters) within the bounding box, and dividing by 10 converts the result to centimeters. Using percentiles instead of extremes ensures that a handful of noisy readings at either end of the distribution do not distort the measurement. If the crop contains fewer than 10 valid pixels, we report the measurement as a sensor error rather than returning a misleading value.

\subsubsection{Segmentation Mask Depth (YOLOv8n-Seg)}

For example, a localized geometric technique is used to establish the street surface baseline in segmentation (YOLOv8n-Seg). The predicted mask is binarized at a $0.5$ threshold after being scaled to the original image dimensions. A buffered circular region is created by stretching the mask's bounding box outward by $15\text{ pixels}$ on each side and deleting the pothole mask itself in order to isolate clean pavement directly surrounding the cavity while avoiding distant structural outliers (such as curbs or walls). In order to remove sensor abnormalities, the street reference depth $d_{\text{street}}$ is calculated as the trimmed median of valid depth values within this annular ring, excluding data below the fifth and above the ninety-fifth percentiles. The final physical depth is then:

\begin{equation}
D_{\text{seg}} = \frac{d_{\text{bottom}} - d_{\text{street}}}{10}
\label{eq:dseg}
\end{equation}

The key advantage of this formulation is geometric. Because the segmentation mask tightly traces the actual edge of the pothole, the annular ring outside it contains only genuine road surface pixels. A bounding box, by contrast, is always rectangular and therefore always encloses some amount of healthy pavement within its boundary.

\begin{figure}[!htbp]
    \centering
    \includegraphics[width=\columnwidth]
    {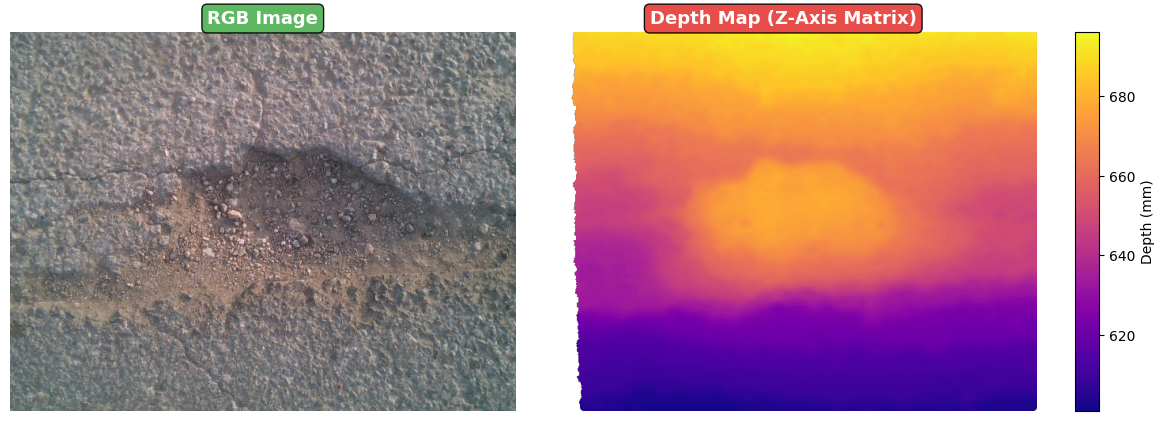}
    \caption{\small A representative RGB-D
    image pair from the PothRGBD
    dataset~\cite{yurdakul2025enhanced}}
    \label{fig:rgbd_pair}
\end{figure}

Fig.~\ref{fig:rgbd_pair} shows a representative RGB-D pair from the PothRGBD dataset.
 
 We measure how much this difference matters in Section~\ref{sec:experiments}, where we run all five models on the same image and compare the extracted depths directly.

\section{Experiments and Results}
\label{sec:experiments}

\subsection{Experimental Setup}

\begin{figure}[!htbp]
    \centering
    \includegraphics[width=\columnwidth]
    {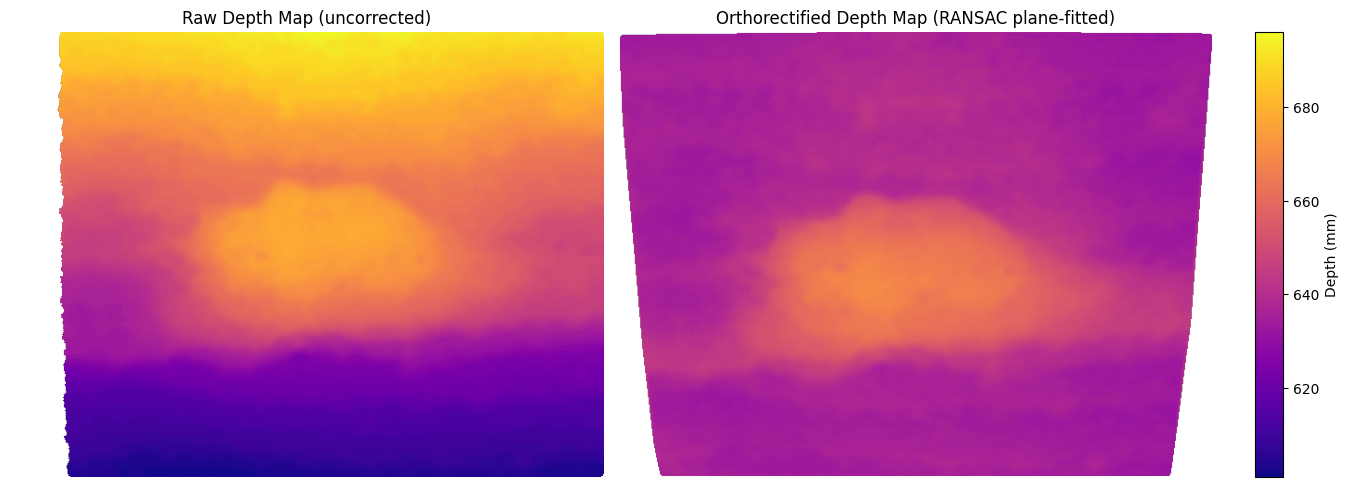}
    \caption{RANSAC orthorectification before testing on all the five architecture.}
    \label{fig:ransac}
\end{figure}

We ran all experiments on a single NVIDIA Tesla T4 GPU as the compute environment. The software stack was Python 3.12.13, PyTorch 2.10.0, CUDA 12.8, and Ultralytics 8.4.46. Each of the five architectures was trained from scratch for 50 epochs at $640 \times 640$ resolution on the augmented training set of 1,600 samples (after augmentation). Then evaluated on the same fixed 200-image validation split that was held out before augmentation was applied. We set the inference confidence threshold to 0.25 for all five models. Prior to depth extraction, all depth maps were orthorectified using the RANSAC procedure in Fig~\ref{fig:ransac}, and all zero-valued D415 pixels were cast to NaN. The exact same RGB and depth image pair was used across all five depth extraction runs.

\subsection{Detection Performance}

Table~\ref{tab:results} reports Precision
(P), Recall (R), mAP@50, and mAP@50-95
for all five architectures on the PothRGBD
validation set.

\begin{table}[htbp]
\caption{Detection Performance on PothRGBD Validation Set (200 Images)}
\label{tab:results}
\centering
\renewcommand{\arraystretch}{1.3}
\begin{tabular}{|l|c|c|c|c|}
\hline
\textbf{Model} & \textbf{P} & \textbf{R} & \textbf{mAP50} & \textbf{mAP50-95} \\
\hline
YOLOv8n     & 0.9408 & 0.8826 & 0.9391 & 0.6281 \\\hline
YOLOv8n-Seg & 0.9151 & \textbf{0.9352} & \textbf{0.9556} & \textbf{0.6758} \\\hline
YOLOv9t     & \textbf{0.9525} & 0.8981 & 0.9529 & 0.6535 \\\hline
RT-DETR-L   & 0.9439 & 0.8704 & 0.9022 & 0.6277 \\\hline
RT-DETR-X   & 0.9431 & 0.9208 & 0.9267 & 0.6281 \\
\hline
\multicolumn{5}{l}{\footnotesize \textbf{Bold} indicates the best value in each column.}
\end{tabular}
\end{table}
With the highest mAP@50 of 0.9556 and the highest mAP@50-95 of 0.6758, YOLOv8n-Seg showed that the segmentation mask head enhances localization accuracy without sacrificing detection quality. With a precision of 0.9525, YOLOv9t had the highest Precision score. In line with the research on Transformer detectors requiring additional data to generate attention representations, RT-DETR-L had the lowest mAP@50 score of 0.9022.

\begin{figure}[htbp]
    \centering
    \includegraphics[width=\linewidth]{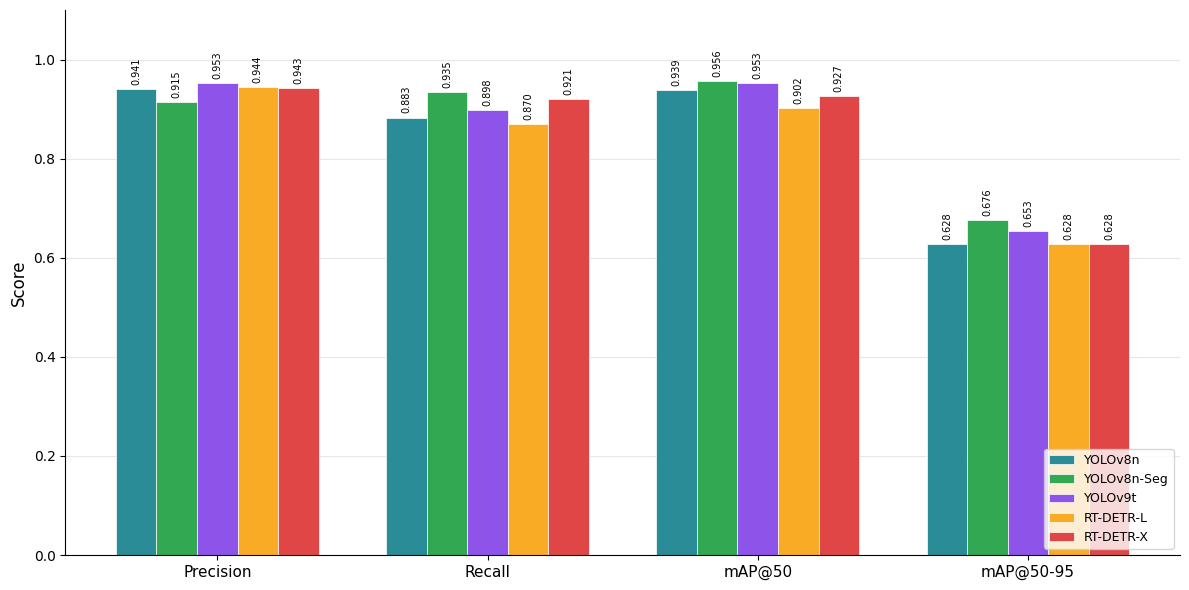}
    \caption{Detection performance across all five architectures on the 200-image PothRGBD validation set.}
    \label{fig:detection_performance}
\end{figure}

Fig. \ref{fig:detection_performance} shows the multi-dimensional trade-offs at a glance by simultaneously visualizing all four performance indicators across the five assessed models.

\subsection{Training Convergence Analysis}

\begin{figure}[h]
    \centering
    \includegraphics[width=0.5\columnwidth]{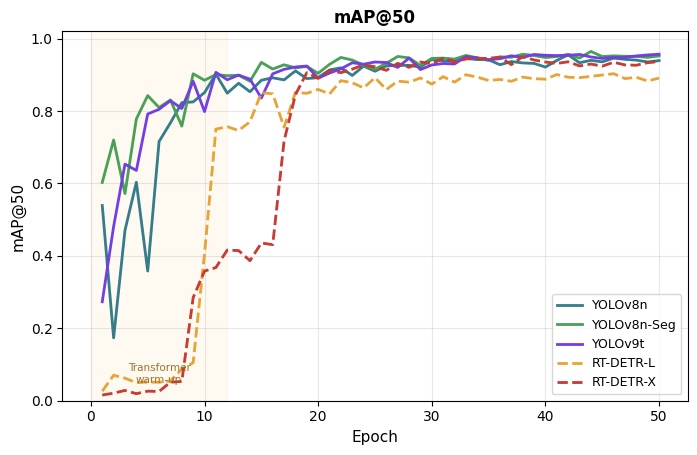}\hfill\includegraphics[width=0.5\columnwidth]{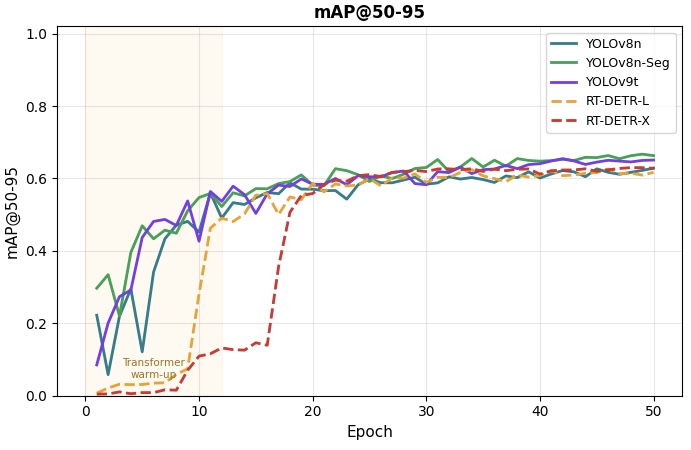}
    \caption{mAP@50 and mAP@50-95 convergence curves for all five architectures over 50 epochs.}
    \label{fig:convergence}
\end{figure}

Fig.~\ref{fig:convergence} shows the mAP@50 and mAP@50-95 convergence curves for all five architectures over 50 training epochs, and the two plots reveal a clear architectural divide in how these models learn.

The three CNN models begin learning from the very first epoch and improve steadily throughout training without any prolonged stall. YOLOv8n-Seg climbs the fastest among them, reaching a mAP@50 of approximately 0.956 by epoch 50. YOLOv9t behaves slightly differently in the early epochs: it shows a noticeable dip at epoch 2 where mAP@50 drops sharply before recovering by epoch 6 and then converging smoothly to approximately 0.953.

Both RT-DETR-L and RT-DETR-X remain mostly flat through the first 12 epochs, shown by the shaded warm-up region in the figure. During this period their mAP@50 values hover near zero, meaning the models are detecting almost nothing useful. Then, around epoch 12 to 13, both models jump sharply as their attention mechanisms finally begin to converge. RT-DETR-L stabilizes at approximately 0.902 by epoch 50. RT-DETR-X follows the same warm-up pattern but converges to a higher level, surpassing RT-DETR-L from approximately epoch 20 onward and reaching 0.927 by epoch 50.

\subsection{Inference Speed Analysis}

Table~\ref{tab:speed} reports inference
latency and model complexity for all five
architectures measured on the Tesla T4 GPU.

\begin{table}[htbp]
\caption{Inference Speed and Model Complexity}
\label{tab:speed}
\centering
\renewcommand{\arraystretch}{1.3}
\begin{tabular}{|l|c|c|c|}
\hline
\textbf{Model} & \textbf{ms} & \textbf{Params (M)} & \textbf{GFLOPs} \\
\hline
YOLOv8n     & \textbf{3.6}  & 3.01  & 8.1   \\\hline
YOLOv8n-Seg & 22.7          & 3.26  & 12.6  \\\hline
YOLOv9t     & 49.5          & 2.00  & 7.9   \\\hline
RT-DETR-L   & 67.3          & 31.99 & 103.4 \\\hline
RT-DETR-X   & 90.2          & 65.47 & 222.5 \\
\hline
\multicolumn{4}{l}{\footnotesize \textbf{Bold} = fastest.}
\end{tabular}
\end{table}

YOLOv8n is the fastest choice, with 3.6 ms and 8.1 GFLOPs. With the mask generation head, YOLOv8n-Seg takes 22.7 ms time. Despite having 2.00 M parameters, YOLOv9t runs at 49.5 ms because of PGI inference overhead. RT-DETR-L runs at 67.3 ms with 103.4 GFLOPs and RT-DETR-X is the slowest at 90.2 ms with 222.5 GFLOPs.

\subsection{RGB-D Physical Depth Measurement Results}

\begin{figure}[!htbp]
    \centering
    \includegraphics[width=0.49\columnwidth]{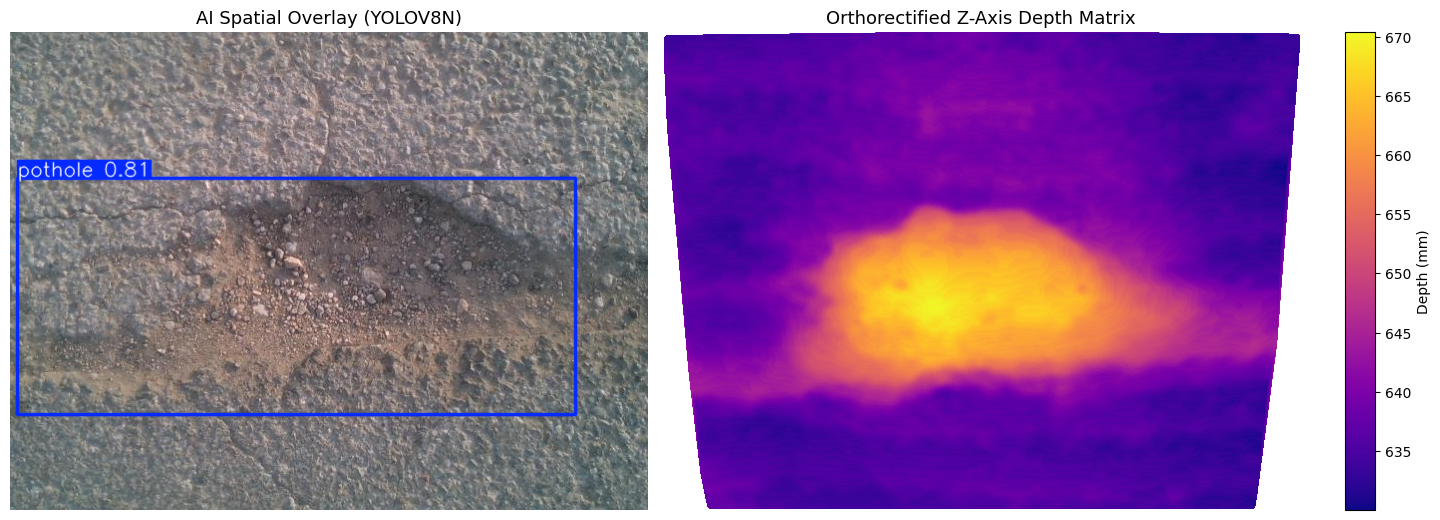}\hfill\includegraphics[width=0.49\columnwidth]{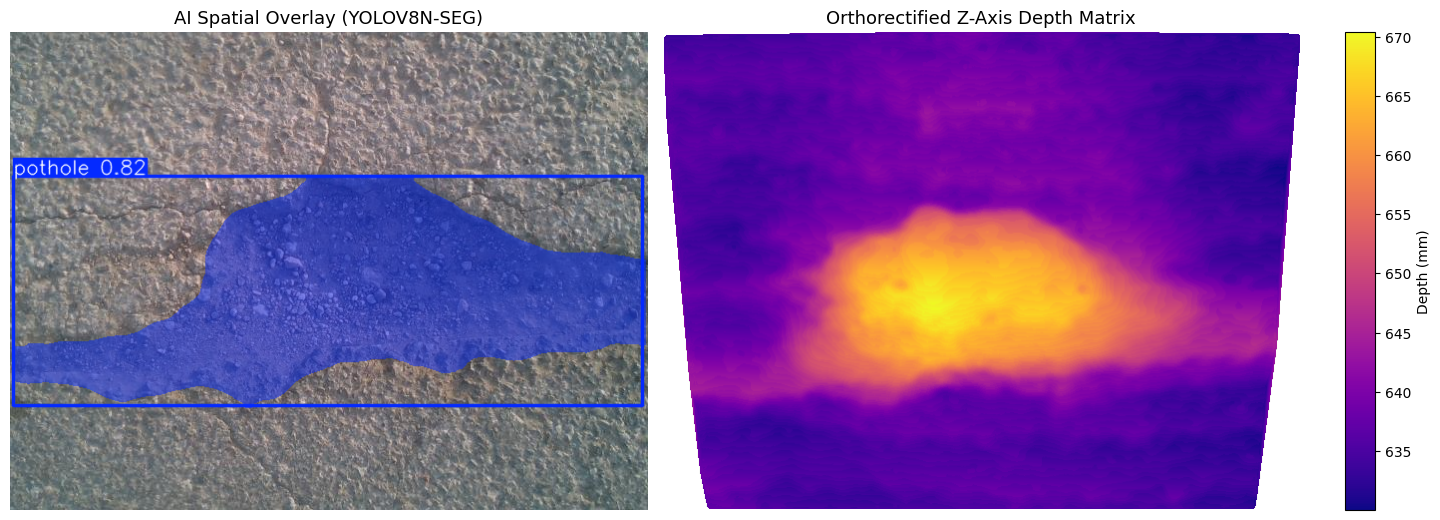}
 
    \vspace{0.5em}
    \includegraphics[width=0.49\columnwidth]{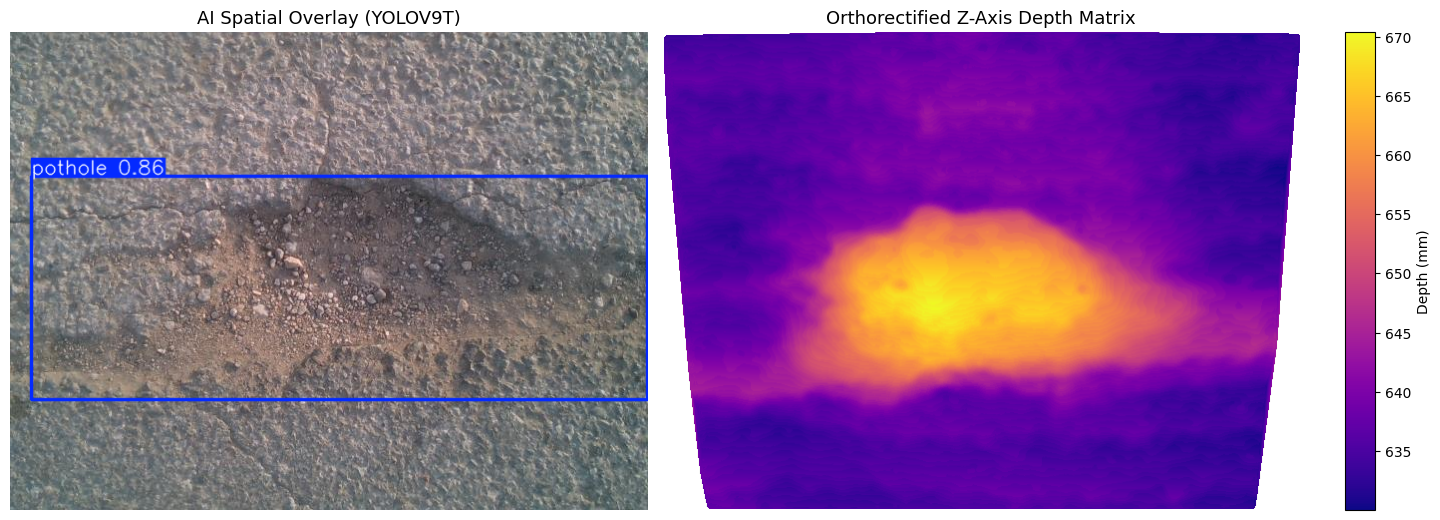}\hfill\includegraphics[width=0.49\columnwidth]{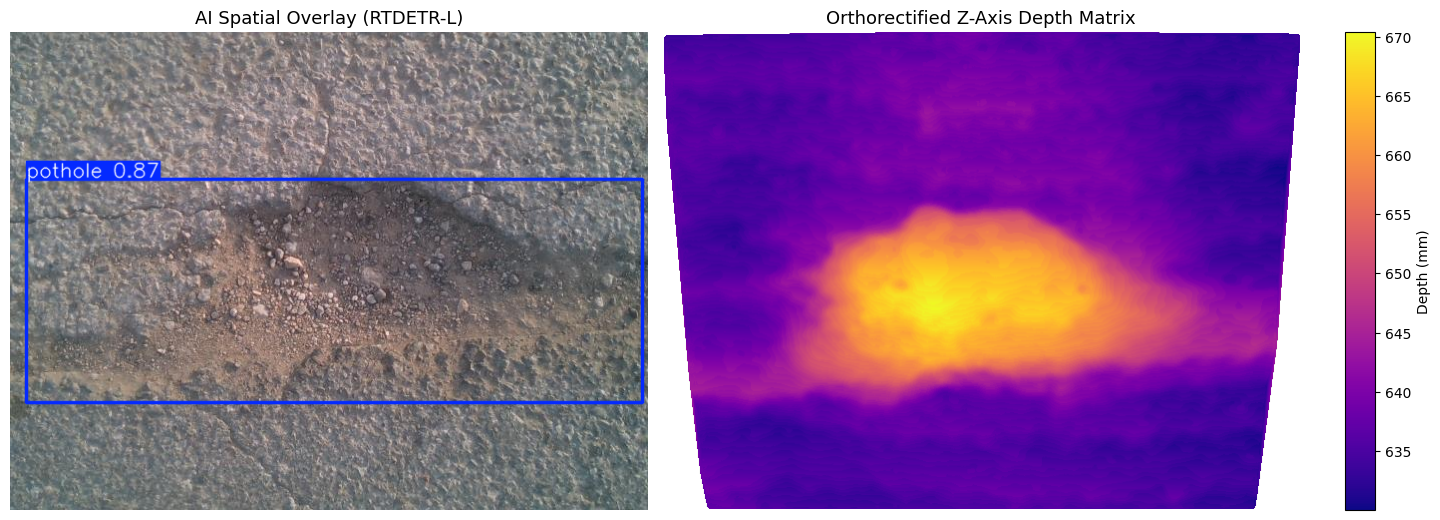}
 
    \vspace{0.5em}
    \includegraphics[width=0.49\columnwidth]{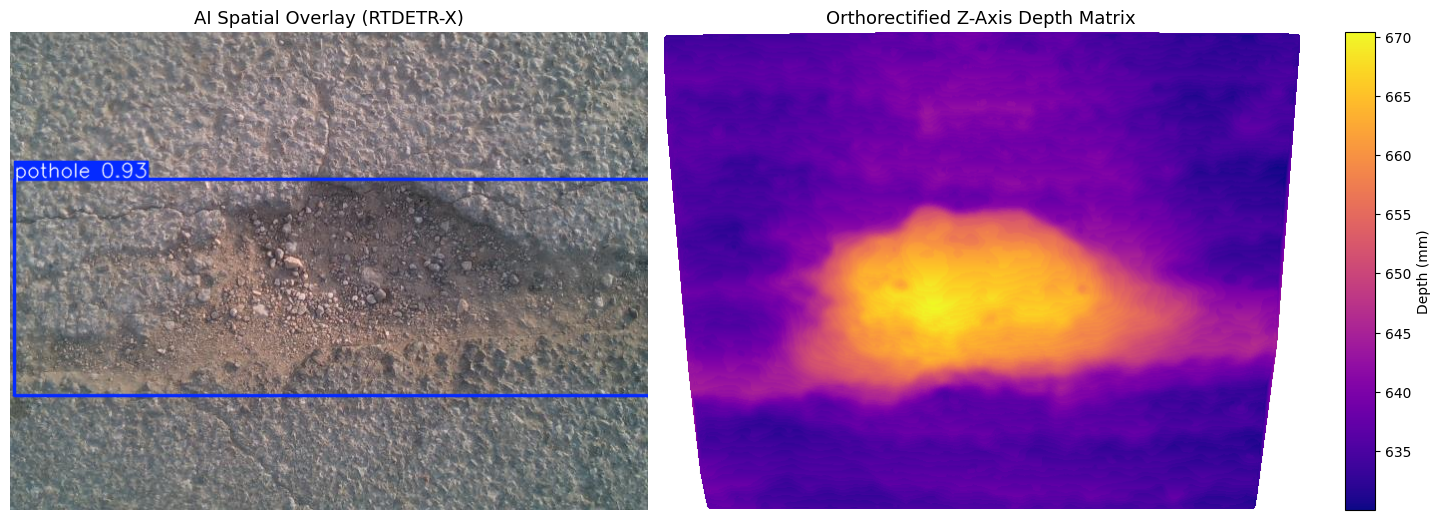}
 
    \caption{\scriptsize RGB-D inference output for all five architectures on the same test image.}
    \label{fig:all_results}
\end{figure}

Table~\ref{tab:depth} shows the depth extraction results for all five
architectures tested on the same matched RGB-depth image pair from the PothRGBD dataset. Prior to all computations are performed, zero-depth pixels are cast to NaN on RANSAC orthorectified depth maps. Fig.~\ref{fig:all_results} shows the
RGB-D inference output for all five
architectures on the same test image.

\begin{table}[htbp]
\caption{RGB-D Physical Depth Extraction Results on the Same Test Image Pair (RANSAC Orthorectified)}
\label{tab:depth}
\centering
\renewcommand{\arraystretch}{1.3}
\begin{tabular}{|l|c|c|c|c|}
\hline
\textbf{Model} & \textbf{Conf.} & \textbf{Output} & \textbf{Depth} & \textbf{Algo.} \\
\hline
YOLOv8n     & 81.12\% & Box  & 3.12 cm & $D_{\text{box}}$ \\\hline
YOLOv8n-Seg & 82.01\% & Mask & \textbf{2.96 cm} & $D_{\text{seg}}$ \\\hline
YOLOv9t     & 85.64\% & Box  & 3.17 cm & $D_{\text{box}}$ \\\hline
RT-DETR-L   & 86.88\% & Box  & 3.16 cm & $D_{\text{box}}$ \\\hline
RT-DETR-X   & \textbf{92.70\%} & Box & 3.16 cm & $D_{\text{box}}$ \\
\hline
\multicolumn{5}{l}{\footnotesize \textbf{Bold} = highest.}
\end{tabular}
\end{table}

RT-DETR-X achieved the highest detection confidence at 92.70\%. This is consistent with the global self-attention mechanism in RT-DETR, which allows the model to reason over the entire image at once rather than relying on local convolutional receptive fields, giving it greater confidence when the visual evidence for a pothole is spread across a large or irregular region.

However, the more important result from this table is not about confidence but about depth. After applying the full RANSAC ground-plane orthorectification, YOLOv8n-Seg extracted a physical depth of 2.96~cm using the $D_{\text{seg}}$ algorithm, while all four bounding box models reported depths between 3.12 and 3.17~cm using $D_{\text{box}}$. That is an overestimate of 0.16--0.21~cm across every single bounding box architecture, regardless of whether the model is a lightweight CNN or a large Transformer.

What makes this finding significant is that the camera tilt has already been fully corrected at this point. The depth map these models are operating on has been reprojected into 3D, rotated to flatten the road surface, and reprojected back to 2D. Any depth gradient introduced by the camera's physical mounting angle has been mathematically removed. Yet the bounding box models still report a higher depth than the segmentation model. This confirms that the overestimation is not a calibration artifact that could be fixed by better sensor alignment --- it is a structural geometric bias inherent to the rectangular shape of bounding boxes. Because a rectangle cannot conform to the irregular boundary of a pothole, it inevitably encloses some amount of healthy surrounding pavement within its edges. Those pavement pixels push the 5th percentile lower than the true cavity floor, inflating the $P_{95} - P_{5}$ spread and producing a depth reading that is systematically too high.

The tight 0.05~cm spread among the four bounding box models (ranging from 3.12~cm for YOLOv8n to 3.17~cm for YOLOv9t) provides further evidence that this is a consistent, systematic bias rather than random measurement noise. If the discrepancy were caused by sensor variability or model instability, we would expect much larger variation across architectures with fundamentally different parameter counts and detection mechanisms. Instead, they all converge to nearly the same overestimate, because they all make the same geometric mistake.

\subsection{Comparative Analysis
and Architectural Trade-offs}

Table~\ref{tab:summary} presents a
consolidated summary of the key
performance trade-offs across all five
architectures.

\begin{table}[htbp]
\caption{Consolidated Performance Summary}
\label{tab:summary}
\centering
\renewcommand{\arraystretch}{1.3}
\begin{tabular}{|l|c|c|c|c|}
\hline
\textbf{Model} & \textbf{mAP@50} & \textbf{ms} & \textbf{Depth} & \textbf{Conf.} \\
\hline
YOLOv8n     & 0.9391 & \textbf{3.6} & 3.12~cm & 81.12\% \\\hline
YOLOv8n-Seg & \textbf{0.9556} & 22.7 & \textbf{2.96~cm} & 82.01\% \\\hline
YOLOv9t     & 0.9529 & 49.5 & 3.17~cm & 85.64\% \\\hline
RT-DETR-L   & 0.9022 & 67.3 & 3.16~cm & 86.88\% \\\hline
RT-DETR-X   & 0.9267 & 90.2 & 3.16~cm & \textbf{92.70\%} \\
\hline
\multicolumn{5}{l}{\footnotesize \textbf{Bold} indicates the best value per column.}
\end{tabular}
\end{table}

No single architecture dominates across all dimensions. YOLOv8n is best for real-time edge deployment at 3.6 ms and 8.1 GFLOPs. YOLOv8n-Seg is the preferred choice when depth accuracy is the primary objective, leading on both mAP@50 (0.9556) and mAP@50-95 (0.6758) while producing the most accurate depth at 2.96 cm via Dseg. YOLOv9t offers the highest Precision (0.9525) with the smallest parameter footprint (2.00 M), making it suitable for safety-critical scanning. RT-DETR-X achieves the highest confidence (92.70\%) and Recall (0.9208), making it best for server-side pipelines where maximizing detection certainty matters more than latency.

\section{Conclusion}
\label{sec:conclusion}

In this paper, we compared five detection architectures used in edge for depth-aware pothole detection on the PothRGBD dataset using a six-technique weather augmentation pipeline. Among all, YOLOv8n-Seg achieved the highest mAP@50 (0.9556) and most accurate depth estimate (2.96 cm). YOLOv8n was fastest at 3.6 ms. RT-DETR-X achieved the highest confidence (92.70\%) and recall (0.9208).

The primary finding is that bounding box models overestimate depth by 0.16–0.21 cm when compared to the segmentation mask approach. This finding formally quantifies the healthy pavement inclusion bias in rectangular detection regions and gives precise recommendations for structure selection in depth-extraction pipelines.

Future work will validate on larger
datasets, deploy on embedded hardware
such as the NVIDIA Jetson, and extend
to multi-class road anomaly detection
including cracking and rutting.

\bibliographystyle{IEEEtran}
\bibliography{conference_101719}

\end{document}